\documentclass[letterpaper, 10 pt, conference]{ieeeconf}  

\IEEEoverridecommandlockouts                              

\usepackage{graphicx}
\usepackage{tabularx} 
\usepackage{multirow}
\usepackage{subcaption}
\usepackage{amsmath,amsfonts}
\usepackage{xcolor}
\usepackage{url}

\title{\LARGE \bf
Comparison of Pedestrian Prediction Models from Trajectory and Appearance Data for Autonomous Driving

}

\author{Anthony Knittel$^{1}$, Morris Antonello$^{1}$, John Redford$^{1}$ and Subramanian Ramamoorthy$^{1, 2}$
\thanks{$^{1}$Applied Research Team, Five AI (Bosch UK), Edinburgh, United Kingdom
        {\tt\small anthony.knittel@five.ai}}%
\thanks{$^{2}$School of Informatics, University of Edinburgh, Edinburgh, United Kingdom}
}%

\begin{document}

\maketitle
\thispagestyle{empty}
\pagestyle{empty}

\begin{abstract}

The ability to anticipate pedestrian motion changes is a critical capability for autonomous vehicles. In urban environments, pedestrians may enter the road area and create a high risk for driving, and it is important to identify these cases.
Typical predictors use the trajectory history to predict future motion, however in cases of motion initiation,  motion in the trajectory may only be clearly visible after a delay, which can result in the pedestrian has entered the road area before an accurate prediction can be made. 
Appearance data includes useful information such as changes of gait, which are early indicators of motion changes, and can inform trajectory prediction. 
This work presents a comparative evaluation of trajectory-only and appearance-based methods for pedestrian prediction, and introduces a new dataset experiment for prediction using appearance. 
We create two trajectory and image datasets based on the combination of image and trajectory sequences from the popular NuScenes dataset, and examine prediction of trajectories using observed appearance to influence futures. 
This shows some advantages over trajectory prediction alone, although problems with the dataset prevent advantages of appearance-based models from being shown. We describe methods for improving the dataset and experiment to allow benefits of appearance-based models to be captured.

\end{abstract}

\section{INTRODUCTION}

Autonomous Vehicles (AV) need to operate in areas where pedestrians are present. Prediction of future behaviour is important for avoiding conflict, especially when vulnerable road users such as pedestrians are present.  
Pedestrian prediction is hard since they can change direction and start or stop moving, and it is high risk, for example if they enter the road area. Conservative estimates of pedestrian motion can allow potential actions to be captured and avoided, however can lead to very conservative driving of an AV and prevent progress.  A better approach is to accurately identify when changes of motion occur, and to use accurate predictions to avoid conflict situations.

\begin{figure}[!t]
    \centering
    \includegraphics[width=0.8\columnwidth]{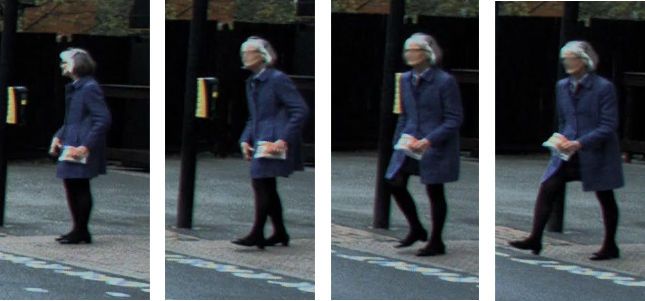}
    \caption{Example of cropped pedestrian appearance, showing gait change with motion initiation}
    \label{fig_problem_example_sequence}
\end{figure}

Existing methods predict future motion based on an observed history of positions.  A significant limitation of these approaches is that when changes of motion take place, such as initiation of motion to enter a road area from a stationary position, there is a delay before the motion can be accurately observed in the trajectory, and used to make an accurate prediction. Noise is present in the estimated position, and the greater the noise the later that motion initiation can be reliably observed.
Appearance cues such as changes of body pose provide additional information about pedestrian actions, such gait changes when pedestrians begin or stop moving.  These appearance cues can reliably inform when motion changes are taking place, and provide an early and accurate signal of motion. Figure~\ref{fig_problem_example_sequence} illustrates an example.

Pedestrian appearance has been used previously to estimate whether a pedestrian intends to cross the road, and to inform prediction of the future position of the pedestrian in the camera view. 
Common datasets are PIE~\cite{Rasouli_2019_ICCV} and JAAD~\cite{rasouli2017, rasouli2017they}. 
These methods demonstrate classification of pedestrian crossing intent, and prediction of future positions within the camera view. 
In order to use these approaches to support an AV, further steps are needed to infer behaviour in the world space in Cartesian coordinates, and it is unclear how well 
camera-based prediction can inform the future world position of pedestrians. 


Prediction of pedestrian motion is inherently a multimodal task- if a pedestrian is standing beside the road area there are at least two significant possibilities to consider, of whether they remain stationary or begin moving into the road. 
A multimodal predictor can create a predicted trajectory for each mode, and assign a probability estimate for each event.  Previous methods~\cite{knittel2022dipa} have described effective methods for evaluating multimodal predictions 
of road users. 
These evaluations test whether distinct modes of behaviour are captured, as well as the probability distribution, which is useful for evaluating multimodal predictions of pedestrian motion.

We present an experimental task for pedestrian prediction, that includes a dataset of cropped images of pedestrians, along with their associated trajectory in world coordinates. This dataset is constructed using data from the NuScenes dataset~\cite{caesar2020NuScenes}, which combines camera information with pedestrian trajectories, to produce an experimental task for pedestrian prediction including the use of observed appearance. This experiment involves using a history of images and trajectory positions, and predicting future positions, evaluated using the multimodal prediction measures from~\cite{knittel2022dipa}.  
This experimental task allows a model to predict behaviour modes, such as motion initiation and standing still, using the appearance of pedestrians to provide cues when changes of motion take place.

To solve this task we compare physics-based models, trajectory-only prediction, and two network architectures using a Convolutional Neural Network (CNN) model and pre-calculated pose features for interpreting pedestrian appearance. 
This examines how pedestrian appearance such as changes of gait, can be used to estimate the future trajectory modes of pedestrians, using a prediction representation that can be used by an AV planner to control the vehicle while avoiding potential conflicts with pedestrians. 

\section{EXISTING METHODS}

A number of approaches have been proposed for prediction of agents in road areas, including physics, goal, and regression methods, using trajectory and appearance data. 

Kinematic models, e.g. constant velocity (CV) or acceleration~\cite{schubert2008comparison}, efficiently capture simple motions and can be a reasonable estimate when an agent is moving consistently. A study~\cite{scholler2020} has suggested that CV models perform as well as data-driven methods for pedestrian trajectories. 
Goal-based methods~\cite{fajen2003behavioral, garzon2016pedestrian, zhao2021tnt, antonello2022flash, mangalam2020not} 
estimate a belief that each goal is being 
pursued by the agent, for example using scene information. 
There can be a large number of possible goals a pedestrian may follow, and it may not be possible to reliably identify goal-directed behaviour.

Regression-ba sed methods directly map observations to predicted outputs. These representations can include interactions between multiple agents of varying classes, 
and map elements. Recent architectures~\cite{zhao2021tnt, gao2020vectornet, knittel2022dipa}
are based on Graph Neural Networks (GNN), which can capture complex representations and interactions. 
Further models have examined estimation of error covariances~\cite{postnikov2021covariancenet}, and multi-modal predictions using Gaussian Mixture Models~\cite{salzmann2020trajectron++}.


Appearance models use images as input, in order to infer the current or future motion of an agent since it allows the pose of a pedestrian to be observed, which provides important cues. 
Some models perform prediction based on a 
fixed elevated camera~\cite{zernetsch2018, carreira2017quo}, although these have limitations for use with AVs which use moving cameras. 
Further models~\cite{rasouli2017they, kotseruba2021benchmark, song2022pedestrian}, focus on intent prediction, e.g. crossing vs not crossing. A disadvantage of these methods is that they require manual intent annotation, which can be hard to define and identify. In contrast, trajectory prediction can be based on observations from sensors and the perception system without requiring additional labeling. Others, e.g.~\cite{Rasouli_2019_ICCV, rasouli2020pedestrian, malla2020titan, rasouli2021bifold}, tackle trajectory prediction in the image space rather than world space.  Each of these methods require further processing stages to be able to infer predicted pedestrian motion in the world space.


\section{PROPOSED EXPERIMENTS}

\subsection{Dataset}

\begin{figure}[!t]
    \centering
    \includegraphics[width=0.8\columnwidth]{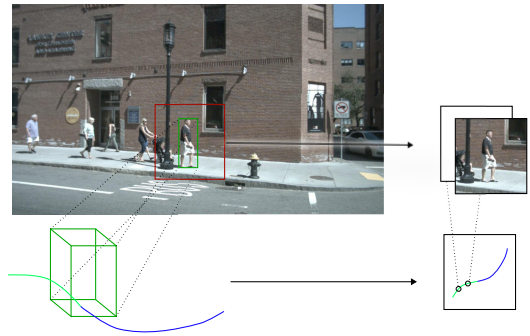}
    \caption{Overview of construction of NuScenes-Appearance dataset. Pedestrian 3D detections are interpolated from 2Hz to 10Hz and projected onto camera frames.  Image crops are recorded in the dataset along with associated trajectory positions for each timestep. Cropped images and trajectory positions are used as input to a predictor, with the objective of predicting the future motion trajectory.}
    \label{fig_dataset}
\end{figure}


To address the task of predicting pedestrian motion from appearance, we construct the NuScenes-Appearance dataset using camera and trajectory information present in the 
NuScenes dataset~\cite{caesar2020NuScenes}. NuScenes contains sensor data collected from a fleet of autonomous vehicles operating in urban environments. It includes 3D trajectory annotations at 2\,Hz and camera images at a variable rate (10 or 20\,Hz). 
We generate the NuScenes-Appearance dataset by interpolating 3D trajectory annotations at 10\,Hz, finding the closest camera timestamp for each camera frame, and projecting the 3D box to form a 2D box in each camera frame.  This box is expanded to a square of twice the largest dimension 
and recorded in an image database, where each cropped image is associated with a recorded trajectory position.

This dataset includes camera images from different views, e.g. front-left, front-right etc cameras. Since each pedestrian can be visible from multiple views, each view is considered a separate trajectory, while individual agents (pedestrians) are kept in the same dataset split. 

We select pedestrian instances from NuScenes and maintain the original data splits. 
As annotations are not provided in the original test set, 
we use the NuScenes validation set as test for the NuScenes-Appearance dataset, and define train and validation sets randomly from the NuScenes train set with a 7:1 ratio.



\subsection{Methods}

We compare the different prediction models on a prediction task with
histories of length 1\,s, and prediction of 3\,s. 
We predict multi-modal trajectories with spatial distributions, and evaluate with standard trajectory error measures: minADE/FDE, predRMS (most probable mode), expRMS (expected RMS) and NLL. These measures evaluate closest-mode prediction as well as probabilistic estimates, which provide complementary evaluations of prediction accuracy~\cite{knittel2022dipa, luo2021safety}. 
An effective predictor needs to perform well on each measure, indicating the ability to capture distinct modes of behaviour, as well as accurate estimates of the probability that each will occur. 

Appearance-based prediction can assist with identifying changes of motion, and to focus on this task we create a dataset selection that emphasises changes of motion, in addition to the full dataset.  Instances with high motion change are defined based on an average displacement error of $>=0.5m$ with a constant-velocity model. The motion-changes dataset is constructed using the instances with high motion change, and an equal number of random selections from the remaining instances.
Predictions are produced with 5 modes, which are encoded using a predicted trajectory position for each timestep, as well as a 2x2 covariance matrix representing the spatial error distribution, and a probability weight for each predicted mode.  Calculation of the evaluation measures minADE/FDE, predRMS and NLL are described in~\cite{knittel2022dipa}, and expRMS in~\cite{luo2021safety}\footnote{We calculate distances based on trajectory positions rather than grid cells as used in \cite{luo2021safety}}. 



%






Experiments are conducted using two appearance-based predictors as described below, 
and a number of trajectory-only predictors, including kinematic prediction (which predicts a single-mode) and a neural-network trajectory predictor (DiPA~\cite{knittel2022dipa}) that has been demonstrated as effective for prediction of road users including pedestrians. 

\section{PROPOSED METHODS}

\begin{figure}[!t]
    \centering
    \includegraphics[width=0.8\columnwidth]{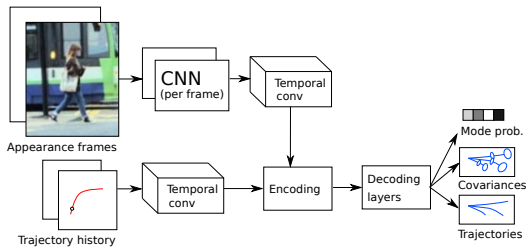}
    \caption{Overview of appearance-based model. Pedestrian appearance of objects is encoded per frame with a CNN and interpreted over time using temporal convolutions.  Image and trajectory encodings are combined and decoded to produce predictions of multimodal trajectories, covariances and mode probabilities to estimate future motion states of pedestrians. 
    } 
    \label{fig_model}
\end{figure}

We describe two appearance-based predictors for utilising an observed sequence of pedestrian images to influence trajectory predictions. The processed images are
combined with the DiPA~\cite{knittel2022dipa} trajectory predictor backbone to predict multimodal future trajectories of predictions.  An overview of the model is shown in Figure~\ref{fig_model}.

One consideration for observing object appearance from the point of view of an autonomous vehicle, is that the camera is moving with the vehicle, and the detected region of each identified pedestrian will contain errors, resulting in visual effects such as background motion and misalignment between sequential frames, which can interfere with processing of visual features.  To compensate for these effects, the appearance-based model processes a sequence of independent image frames using image features (two-dimensional), without the use of temporal video features (three dimensional including time).  
This is followed by temporal convolutions to provide inference between frames over time.
The encoded features representing appearance are concatenated with the trajectory encoding features, and fed into the trajectory decoder of the DiPA model~\cite{knittel2022dipa}. 
We test two implementations, one (App-net) uses a CNN (MobileNetV3Small~\cite{howard2019searching}) which is trained against the mode prediction loss, and a second implementation (App-pose) using pre-calculated pose features~\cite{bajpai2021movenet} which are passed to the temporal convolution layer as a vector of $17\times2$ features of pose positions in the image.
These appearance-based predictors allow visual cues to influence predicted trajectories, and the estimates of probabilities of each trajectory mode. 

The DiPA model used for the trajectory prediction experiments uses the same network backbone, without using the feed from the appearance model.  The DiPA predictor uses stages of temporal convolution, and MLP layers for processing the encoding and decoding outputs of mode probabilities, covariances and trajectory modes.  The original model processes interactions between neighbouring agents, however as this experiment performs single agent prediction agent-agent interactions are not used.  Training is performed using the losses described in~\cite{knittel2022dipa}, which balance capturing distinct behaviour modes with estimating the probability distribution accurately.

\section{RESULTS}

\begin{table}[!t]
\caption{Comparison on the NuScenes-Appearance dataset.}
\label{NuScenes-comparison}
\begin{center}
\resizebox{\columnwidth}{!}{%
\begin{tabular}{c c | c c c | c c c}
\hline
\multicolumn{2}{c|}{\multirow{1}{*}{Dataset}}& \multicolumn{3}{c|}{Full} & \multicolumn{3}{c}{Motion Changes} \\
\cline{1-8}
\multicolumn{2}{c|}{\multirow{1}{*}{Time Horizon}}& 1 s & 2s & 3s & 1 s & 2s & 3s \\
\hline
\multicolumn{1}{c|}{\multirow{2}{*}{RMS}}  & CV & 0.20 & 0.49 & 0.84 & 0.36 & 0.88 & 1.47\\
\multicolumn{1}{c|}{}                          & DA & 0.25 & 0.59 & 0.98 & 0.40 & 0.98 & 1.61\\
\hline
\multicolumn{1}{c|}{\multirow{3}{*}{predRMS}}  & DiPA~\cite{knittel2022dipa} & \textbf{0.18} & \textbf{0.42} & \textbf{0.73} & 0.31 & \textbf{0.75} & \textbf{1.25}\\
\multicolumn{1}{c|}{}                          & \mbox{App-net} & \textbf{0.18} & 0.44 & 0.75 & 0.31 & 0.76 & 1.27\\
\multicolumn{1}{c|}{}                          & \mbox{App-pose} & 0.19 & 0.47 & 0.80 & \textbf{0.30} & \textbf{0.75} & 1.26\\
\hline
\multicolumn{1}{c|}{\multirow{3}{*}{minADE}}    & DiPA~\cite{knittel2022dipa} & \textbf{0.03} & \textbf{0.08} & \textbf{0.13} & \textbf{0.06} & \textbf{0.14} & \textbf{0.23}\\
\multicolumn{1}{c|}{}                           & \mbox{App-net} & 0.04 & \textbf{0.08} & 0.14 & 0.07 & 0.15 & 0.24\\
\multicolumn{1}{c|}{}                           & \mbox{App-pose} & 0.05 & 0.12 & 0.20 & 0.09 & 0.20 & 0.34\\
\hline
\multicolumn{1}{c|}{\multirow{3}{*}{minFDE}}    & DiPA~\cite{knittel2022dipa} & \textbf{0.06} & \textbf{0.16} & \textbf{0.28} & \textbf{0.12} & \textbf{0.28} & \textbf{0.48}\\
\multicolumn{1}{c|}{}                           & \mbox{App-net} & 0.07 & \textbf{0.16} & 0.29 & \textbf{0.12} & 0.29 & 0.49\\
\multicolumn{1}{c|}{}                           & \mbox{App-pose} & 0.10 & 0.25 & 0.43 & 0.18 & 0.44 & 0.74\\
\hline
\multicolumn{1}{c|}{\multirow{3}{*}{expRMS}} & DiPA~\cite{knittel2022dipa} & 0.20 & 0.47 & \textbf{0.73} & 0.35 & 0.87 & 1.46\\
\multicolumn{1}{c|}{}                           & \mbox{App-net} & 0.20 & 0.48 & 0.81 & 0.36 & 0.87 & 1.45\\
\multicolumn{1}{c|}{}                           & \mbox{App-pose} & \textbf{0.19} & \textbf{0.46} & 0.80 & \textbf{0.30} & \textbf{0.75} & \textbf{1.28}\\
\hline
\multicolumn{1}{c|}{\multirow{3}{*}{NLL}} & DiPA~\cite{knittel2022dipa} & \textbf{-1.78} & \textbf{-0.43} & \textbf{0.58} & \textbf{-0.83} & \textbf{0.86} & \textbf{1.79}\\
\multicolumn{1}{c|}{}                     & \mbox{App-net} & -1.75 & -0.38 & 0.64 & -0.79 & 0.88 & 1.82\\
\multicolumn{1}{c|}{}                     & \mbox{App-pose} & -1.15 & 4.96 & 15.90 & 1.34 & 16.19 & 26.81\\
\hline
\end{tabular}}
\end{center}
\end{table}

We compare methods on the two presented datasets using standard trajectory error metrics. Baselines include a Constant Velocity (CV) and a Decaying Acceleration (DA) model. DA relies on Constant Acceleration for short-term and Constant Velocity for the long-term using an exponential decay function, $a_o e^{-\lambda t}$, where $a_0$ is the initial observed acceleration and the decay rate $\lambda$ equals to $5.5\,s^{-1}$. Results are reported in Table~\ref{NuScenes-comparison}. Among physics-based models, CV is best. Accelerations can capture motion initiations, but higher order derivatives are more difficult to estimate and noisy values can be detrimental.  
Since unimodal and multimodal prediction are distinct tasks, unimodal predictors are evaluated with RMS only, which is comparable to predRMS. We do not evaluate unimodal physics-based models with other metrics accounting for multi-modality and uncertainty.

The DiPA trajectory-only prediction model provides accurate predictions that improves over the physics-based baselines, and captures distinct behaviours along with good probabilistic estimates, on both the full and motion-changes datasets. Differences between results on the full dataset are small, as the data is dominated by simple motion behaviours, which does not allow differences between models on capturing changes of motion to be seen.

The App-pose model improves over other models on expRMS but shows higher error on NLL and minADE/FDE.  This indicates that the model has learnt to accurately capture which mode is more likely, however has also followed a conservative mode generation policy that results in instances of the dataset not being covered by the model.  The App-net model produces balanced predictions on the various evaluation measures, although has not shown advantages over the other models. 

These results show the benefits of multimodal evaluations for describing different aspects of how well predictions capture observed behaviours, which is useful for 
pedestrian trajectory prediction. Some advantages of appearance-based models can be seen, however further development is needed to allow appearance cues to provide substantial advantages over trajectory-based predictors.

Analysis of the data shows some significant limitations of the data and experiment.  In a number of cases where a pedestrian 
initiates motion, movement in the trajectory is observed before motion or changes of gait occur according to the observed images.  This effect originates in the source data, which may be caused by retrospective smoothing that allows future positions to influence earlier trajectory positions.  A further effect occurs 
from interpolation between trajectory samples when upsampling detections from 2Hz to 10Hz.  
This effect provides unreasonable advantages to trajectory prediction, allowing trajectory-only prediction methods to be aware of motion before it takes place, and preventing 
advantages of appearance-based models of being demonstrated.  This issue can be addressed through the use of a dataset with a higher sampling rate of observations, and ensuring that dataset filtering does not allow future information to influence earlier timesteps.  
A further issue 
is that in many instances the ground-truth motion does not accurately describe 
the pedestrian motion as observed in the video, for example showing trajectory motion where a person is standing still.  These errors in data will introduce incorrect measurements of performance, for example a confident prediction (with narrow covariance) of stationary motion will be heavily penalised with high errors on NLL scores. 
Higher annotation accuracy would allow the advantages of appearance-based prediction to be more accurately measured.


%
\begin{figure}[!t]
\begin{subfigure}{.5\textwidth}
  \centering
  \includegraphics[width=0.8\columnwidth]{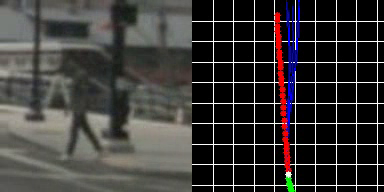}
  \label{fig_qualitative_example_b}
\end{subfigure}
\begin{subfigure}{.5\textwidth}
  \centering
  \includegraphics[width=0.8\columnwidth]{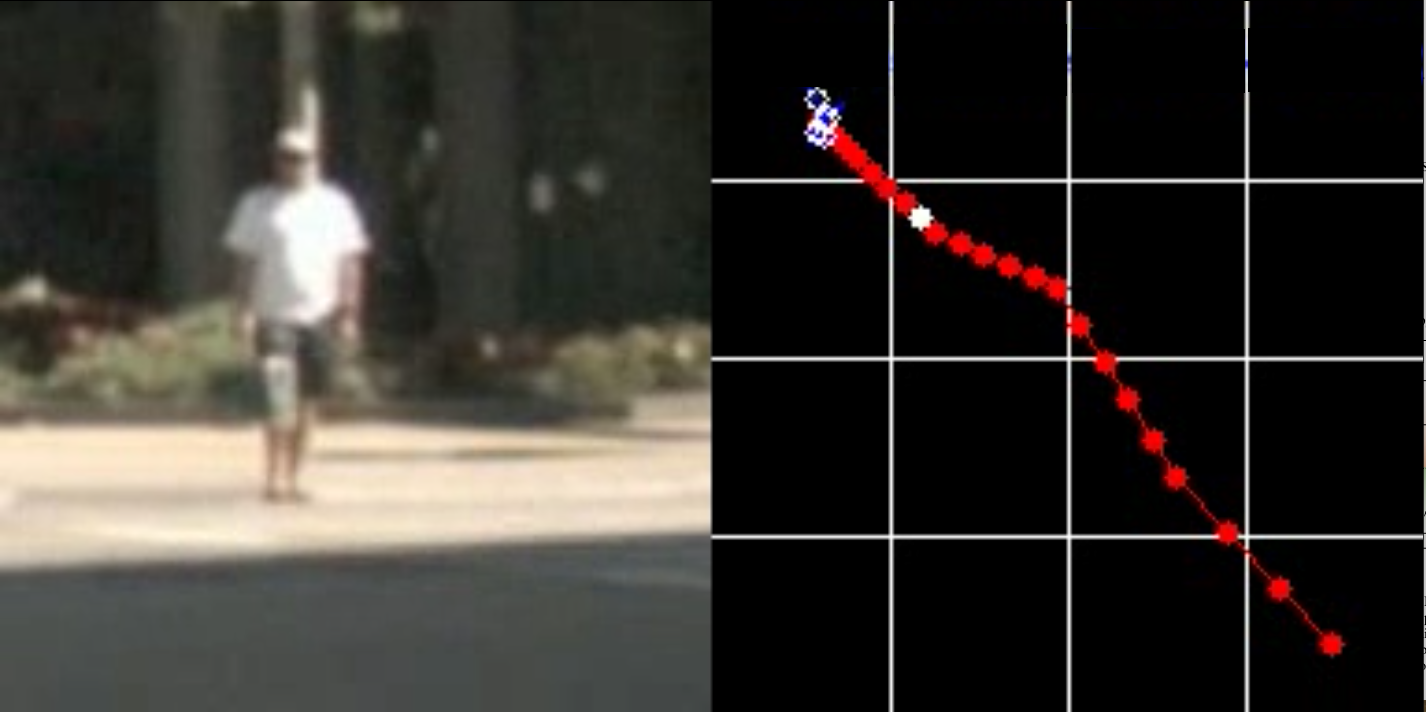}
  \label{fig_qualitative_example_c}
\end{subfigure}

\caption{Examples of observed 
appearance data, along with ground-truth (past - green, future - red, white - prediction point) and multimodal predicted (blue) trajectory data. Top: successful case of motion initiation prediction. Bottom: example demonstrating limitations of source data. Ground-truth trajectory (red) shows motion while pedestrian is still stationary, and pre-empts motion before it occurs, for example due to bidirectional filtering over time. 
}
\label{fig_qualitative_example}
\end{figure}




\section{SUMMARY}

In order to operate an autonomous vehicle in the vicinity of pedestrians, it is important to be able to estimate their future motion, and to identify significant cases such as changes of motion, which can indicate when they may enter the road area.  To address 
this problem we introduce a new dataset task to perform estimation of multimodal trajectories, using pedestrian appearance to inform future motion.  
This task improves over previous datasets such as PIE and JAAD, which are limited to the camera frame, by evaluating prediction of motion in the world space, and including evaluation of probabilistic estimates of different modes of motion.

Comparison of these models shows that the neural-network trajectory predictor improves over the kinematic model and provides accurate predictions on all evaluation measures. The pose-based model improves on weighted trajectory estimates, indicating accurate mode estimation, however shows higher error on other tasks as a result of a conservative mode estimation strategy.  
Appearance-based prediction can provide advantages from using motion cues to inform predicted trajectories, however further development on this topic is needed to clearly capture these advantages.


An important limitation of the dataset, is that trajectory samples include motion before it takes place, 
for example as a result of filtering of the dataset and through interpolation.  
These effects interfere with the advantages of appearance-based prediction from being demonstrated.
An improved experiment can be made by ensuring that dataset filtering does not allow future information to influence earlier timesteps, which will provide a more realistic experiment that corresponds with real-world usage.
A further limitation 
is that it operates for a single pedestrian at a time,  
while further improvements can support the prediction of multiple agents together in a scene, including the use of appearance for each agent.  

\bibliographystyle{unsrt}
\bibliography{bibliography}

\end{document}